\documentclass[10pt,twocolumn,letterpaper]{article}

\usepackage{cvpr}
\usepackage{times}
\usepackage{epsfig}
\usepackage{graphicx}
\usepackage{amsmath}
\usepackage{amssymb}

\usepackage{url}
\usepackage{booktabs}
\usepackage{multirow}
\usepackage{colortbl}
\usepackage{tabu}
\usepackage{pifont}

\usepackage[breaklinks=true,bookmarks=false]{hyperref}

\cvprfinalcopy 

\def\cvprPaperID{****} 

\begin{document}

\title{3rd Place: A Global and Local Dual Retrieval Solution to Facebook AI Image Similarity Challenge}

\author{Xinlong Sun$^*$, Yangyang Qin\thanks{Equal contribution} , Xuyuan Xu, Guoping Gong, Yang Fang, Yexin Wang\\
AI Technology Center, Online-Video BU, Tencent\\
Shenzhen, China\\
{\tt\small \{xinlongsun,joeqin,evanxyxu,guopinggong,yangelfang,yexinwang\}@tencent.com}
}

\maketitle

\begin{abstract}
  As a basic task of computer vision, image similarity retrieval is facing the challenge of large-scale data and image copy attacks. This paper presents our 3rd place solution to the matching track of Image Similarity Challenge (ISC) 2021 organized by Facebook AI. We propose a multi-branch retrieval method of combining global descriptors and local descriptors to cover all attack cases. Specifically, we  attempt many strategies to optimize global descriptors, including abundant data augmentations, self-supervised learning with a single Transformer model, overlay detection preprocessing. Moreover, we introduce the robust SIFT feature and GPU Faiss for local retrieval which makes up for the shortcomings of the global retrieval. Finally, KNN-matching algorithm is used to judge the match and merge scores. We show some ablation experiments of our method, which reveals the complementary advantages of global and local features.

\end{abstract}

\section{Introduction}

Image retrieval is a fundamental task of computer vision and multimedia processing. It is to find similar images from reference images for query images. Depending on different tasks, the definition of similarity will be slightly different. In recent years, large-scale retrieval has become more and more important and practical. Different from the Google Landmarks Datasets \cite{weyand2020google} composed of natural landmark pictures, the Facebook AI Image Similarity Challenge\footnote{\href{https://www.drivendata.co/blog/image-similarity-challenge/}{https://www.drivendata.co/blog/image-similarity-challenge/}} provided a new challenging datasets originating from social media platforms, which is a new benchmark for large-scale image similarity detection \cite{douze20212021}. 

\begin{figure}
\begin{center}
\includegraphics[width=3.2in]{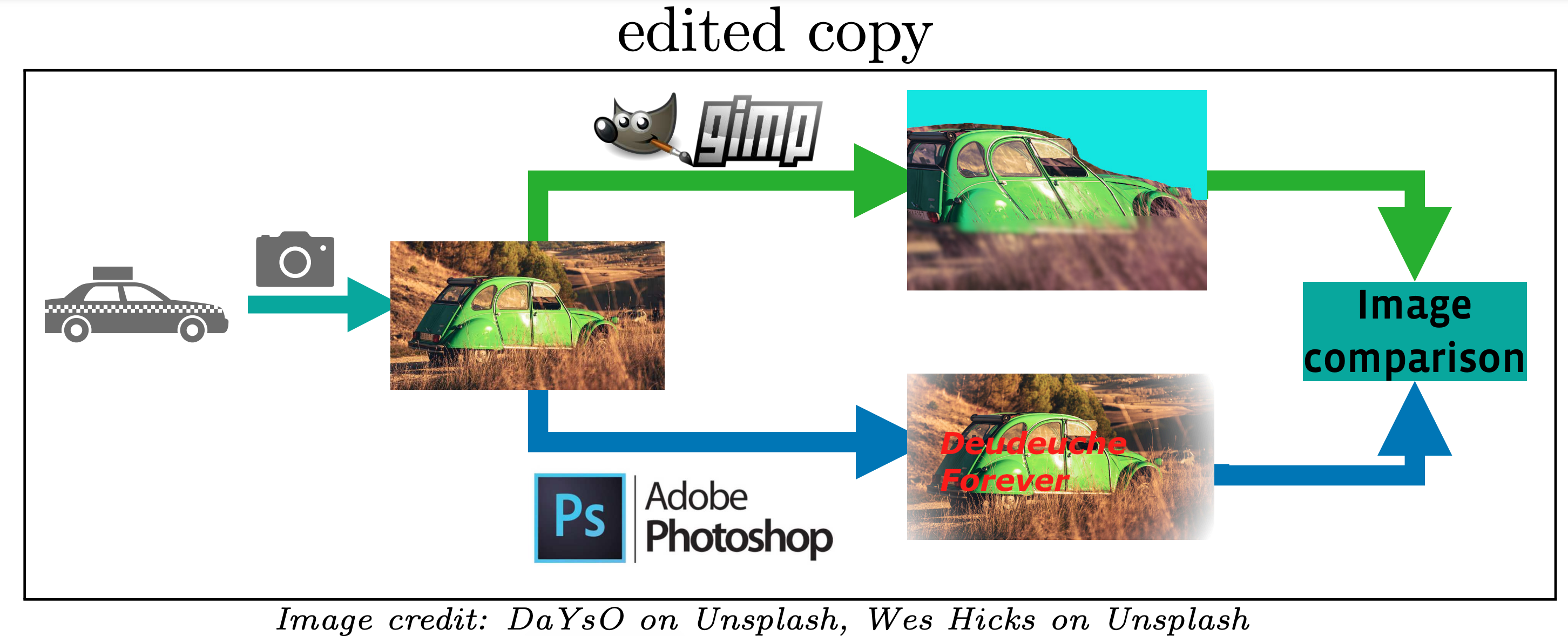}
\end{center}
  \caption{Competition introduction illustration from \cite{douze20212021}.}
\label{fig:competition}
\end{figure}

This competition datasets is mainly suitable for manipulative advertising, preventing uploads of graphic violence, copy detection and enforcing copyright protections \cite{douze20212021}. The goal is to determine whether a query image is a modified copy of any image in a reference corpus of size 1 million, as shown in Fig. \ref{fig:competition}. And the main challenge is that query images will be imposed a lot of attacks during evaluation,  such as crop, blur, rotate, flip, color transformations, spatial transformations, manual manipulation and so on. 

In our solution, we focus on improving the robustness of retrieval for the Facebook AI Image Similarity Challenge. Specifically, we propose a retrieval method of combining global descriptors and local descriptors to cover all attack cases. And we tried many strategies to further optimize, including self-supervised learning, memory bank, overlay detection and multi-branch retrieval. What's more interesting is that our method is very robust, and our model settings only use a single network and traditional SIFT \cite{lowe1999object} feature extraction, which is valuable for online practical applications.

\section{Method}
In this section, we present the overall design and some details of our solution. We firstly introduce our data augmentation methods and  provide the details of preprocessing module. And then we describe how to  get global descriptors and local descriptors. Finally, we describe the image recall pipline and how to rerank images recalled.

\begin{figure*}
\begin{center}
\includegraphics[width=6.4in]{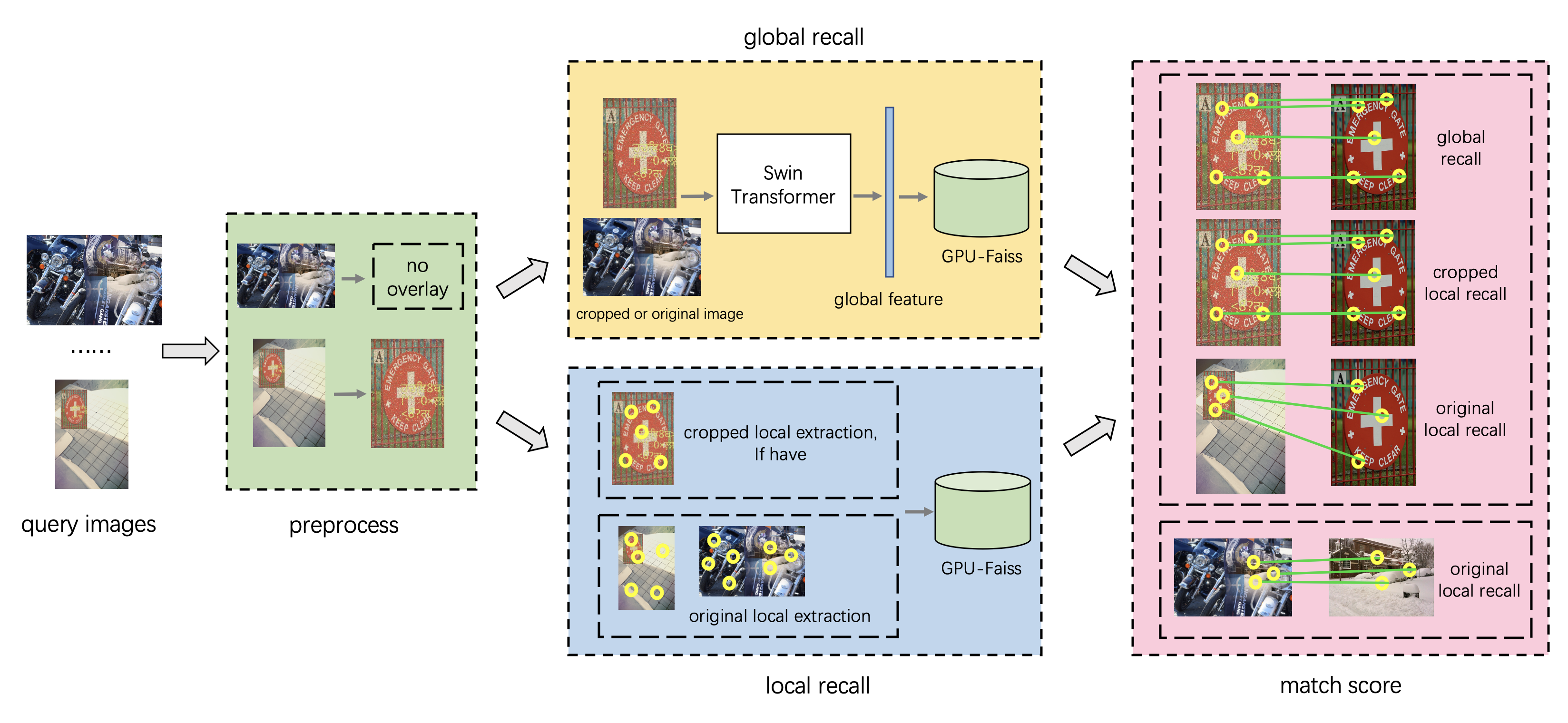}
\end{center}
  \caption{Overall of our dual retrieval solution. }
\label{fig:overall}
\end{figure*}

\subsection{Overall}

  As shown in Fig. \ref{fig:overall}, our pipeline is divided into four modules. When using an image for query, it is first put into the preprocessing module for overlay detection. Then the global and local features are extracted and retrieved in parallel. There are three recall branches: global recall, original local recall and  cropped local recall. The last module will compute the matching score of three branches and merge them into the final result. Specific details will be described below.

\subsection{Datasets and Data Augmentation}
\label{sec:augmentation}
The reference database consists of the set of the original images, while the query set are transformed into the edited copies from reference database or distractors. Moreover, the training datasets, which is of similar spirit to the reference set,  is mainly used to train the model. It is noteworthy that training datasets have no labels, which inspires us to do self-supervised learning. And we use large amounts of data augmentation (more than 40) to generate positive sample pairs to cover most machine-generated attacks, which mainly use some OpenCV and Augly \cite{bitton2021augly} image processing libraries. Meanwhile, different data sources will be treated as negative samples.

\subsection{Preprocessing Module}

\begin{figure}
\begin{center}
\includegraphics[width=3.2in]{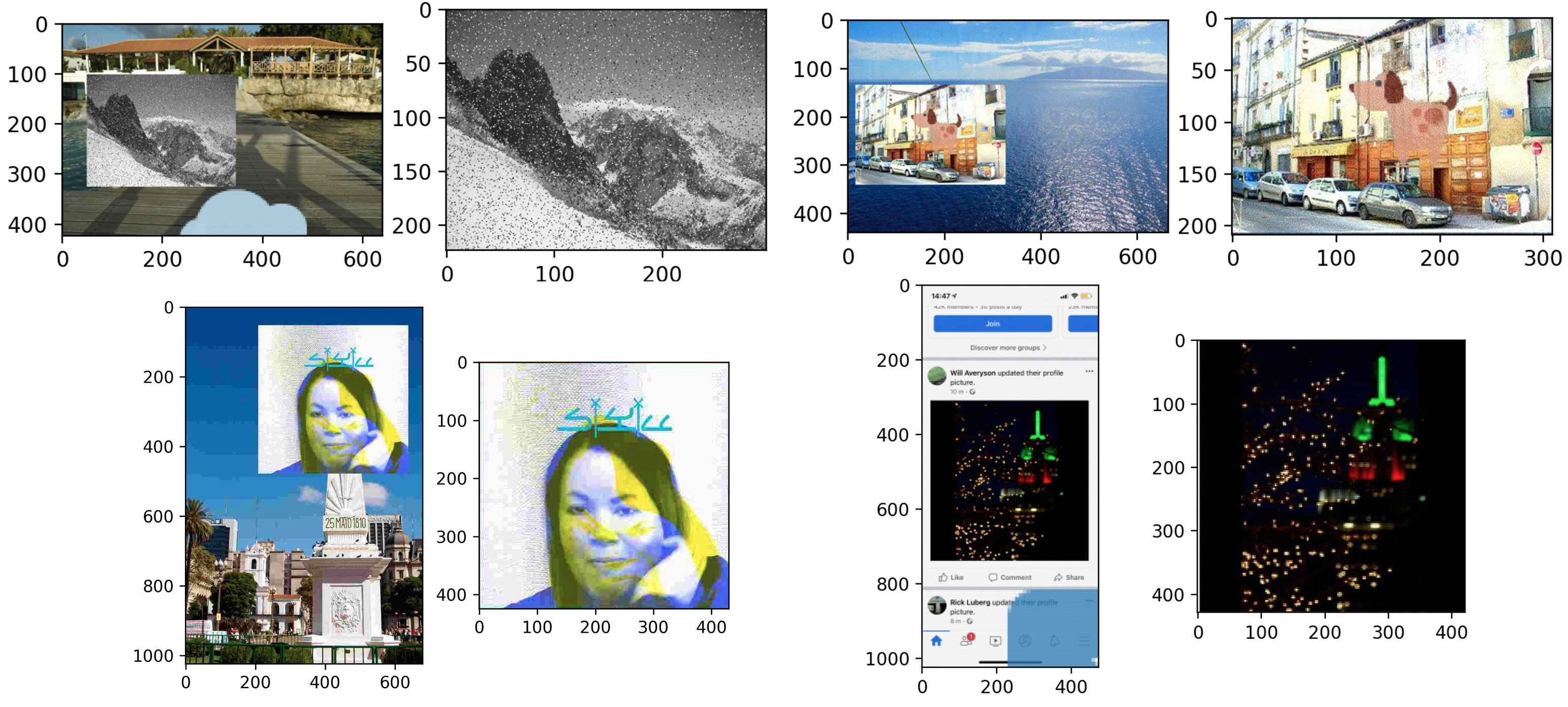}
\end{center}
  \caption{Detection results of our Preprocessing Module. The left is the original image, and the right is the cropped foreground image in each image pair.}
\label{fig:tietu_sample}
\end{figure}

\begin{figure*}
\begin{center}
\includegraphics[width=6.5in]{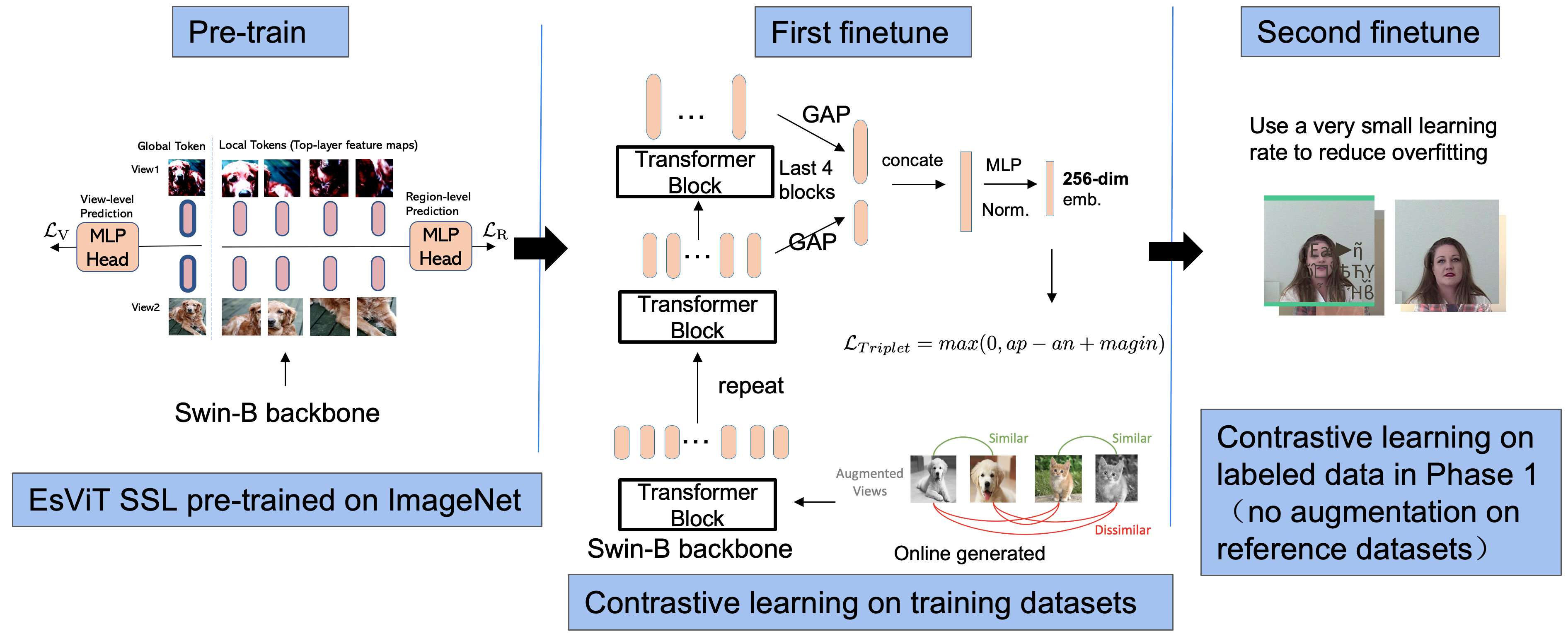}
\end{center}
  \caption{Multi-step training for global descriptor.}
\label{fig:retrieval_sample}
\end{figure*}

First of all,  we trained a pasted image detector as a preprocessing module to detect the foreground image pasted on an unrelated background image, as a preprocessing module against the overlay attacks on query images. This can greatly optimize the image pair similarity of overlay attack. 

To prepare the detection datasets,  we apply Augly \cite{bitton2021augly} overlay function processing on the training datasets to generate pasted images and save the bounding-box labels. Besides, we found that there are a lot of  text and emoji overlaid on the query image, which would cause false positive predictions and affect the detection of pasted foreground images. Therefore, we make the corresponding augmentations to the generated detection datasets to improve the robustness of preventing misjudgments. For detection model, we use an out-of-the-box YOLOv5\footnote{\href{https://github.com/ultralytics/yolov5}{https://github.com/ultralytics/yolov5}} model. The sample of detection images can be seen in Fig. \ref{fig:tietu_sample}.

If a query image have the detection result, the global branch will only use the cropped part to extract feature, while the local branch will both use original and cropped image. This has brought an improvement of one to two percentage points to our overall effect.

\subsection{Global Descriptor}
Recently, self-supervised learning (SSL) with Transformers \cite{caron2021emerging, li2021esvit} enjoyed the same success in CV. SSL can learn general-purpose visual representation without relying on manual supervisions, which is particularly suitable for this large-scale similar retrieval task. 

\begin{figure*}
\begin{center}
\includegraphics[width=6in]{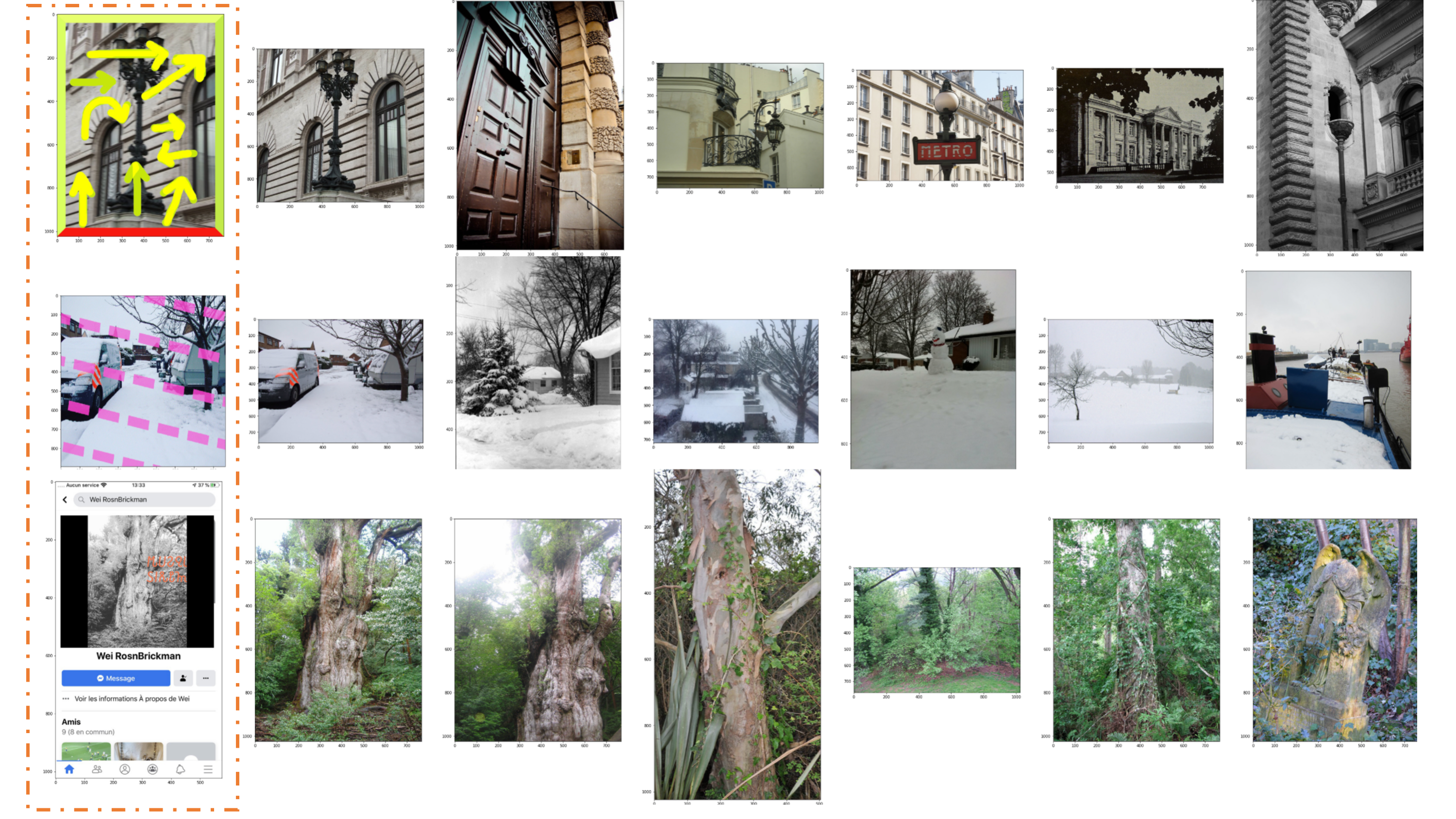}
\end{center}
  \caption{Visualization of global retrieval results. The images in the leftmost dashed box is the query image of Phase2, and the right is the top retrieval results in the reference database.}
\label{fig:retrieval_sample}
\end{figure*}

~

\noindent\textbf{Model Training.}
Based on these observations, we constructe an EsViT SSL model \cite{li2021esvit} with the Swin-B Transformer \cite{liu2021Swin} backbone, which has better generalization performance. As a self-supervised learning model, EsViT uses both region-level and view-level matching pre-train task, which further improves the learned representations and attentions. The details of training process on ImageNet with no lables can refer to \cite{li2021esvit}.

Second, we perform global average pooling operation on the last 4 block feature maps of the Swin-B backbone and get [512, 512, 1024, 1024] dimensional features. We concate them and use a fully connection layer to generate 256-dimensional global descriptor/embedding for each image of similar retrieval. And then perform self-supervised contrastive learning on training datasets for first finetune. We follows the usual image retrieval method to conduct contrastive learning and training by constructing triples. The generation of positive and negative sample pairs of triplet loss training relies on data augmentation as described in Section \ref{sec:augmentation} and the section of Loss Function.

Finally, we use a very small learning rate to perform second finetune on a limited number of labeled data in Phase 1, which also construct triplet loss for contrastive learning, including the query and the reference datasets (without data augmentation, train for distancing the negative samples of reference) . 
See our code for more details. 
 And we can get the final 256-dimensional global descriptor. 
 
 For practical considerations, model ensemble is not used. In addition to the final choice of EsViT, we also tried CNN and DINO \cite{caron2021emerging} models. The model optimization results can be seen in Table 1.

~

\noindent\textbf{Loss Function.}
For triplet loss \cite{schroff2015facenet}, we take images with different ids as negative sample and augmented images from same data sources as positive sample. And we use Euclidean distance to measure the distance between features, and get anchor-positive distance $ap$ and anchor-negative distance $an$. Next, conduct Triplet loss as:

$$
\mathcal{L}_{Triplet} = max(0, ap - an + magin)
$$
Not just using the hardest triplet samples, we fully dig out both hard triplets and semi-hard triplets through comprehensive weighing anchor-positive pair distance and  anchor-negative pair distance.
In addition to the traditional triplet loss, we use XBM \cite{wang2020cross} triplet loss to fully mine negative samples. In detail, we build a memory bank to store historical features of each batch in the past, and use queues for continuous update.

\begin{figure}
\begin{center}
\includegraphics[width=3.2in]{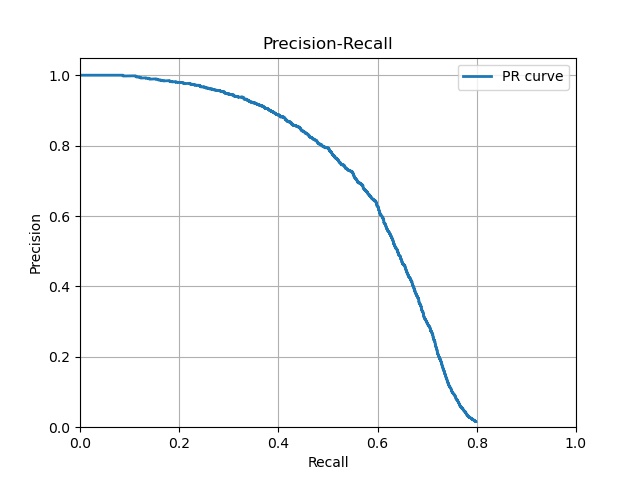}
\end{center}
  \caption{PR curve for only using global retrieval.}
\label{fig:AP_curve}
\end{figure}

~

\noindent\textbf{Visualization.}
We make visualization on global retrieval results in Fig. \ref{fig:retrieval_sample}, and PR curve in Fig. \ref{fig:AP_curve}. Note the PR curve is global model only trained on training datasets, not finetuned on Phase1 labeled data.

\subsection{Local Descriptor}

\begin{figure*}
\begin{center}
\includegraphics[width=6.0in]{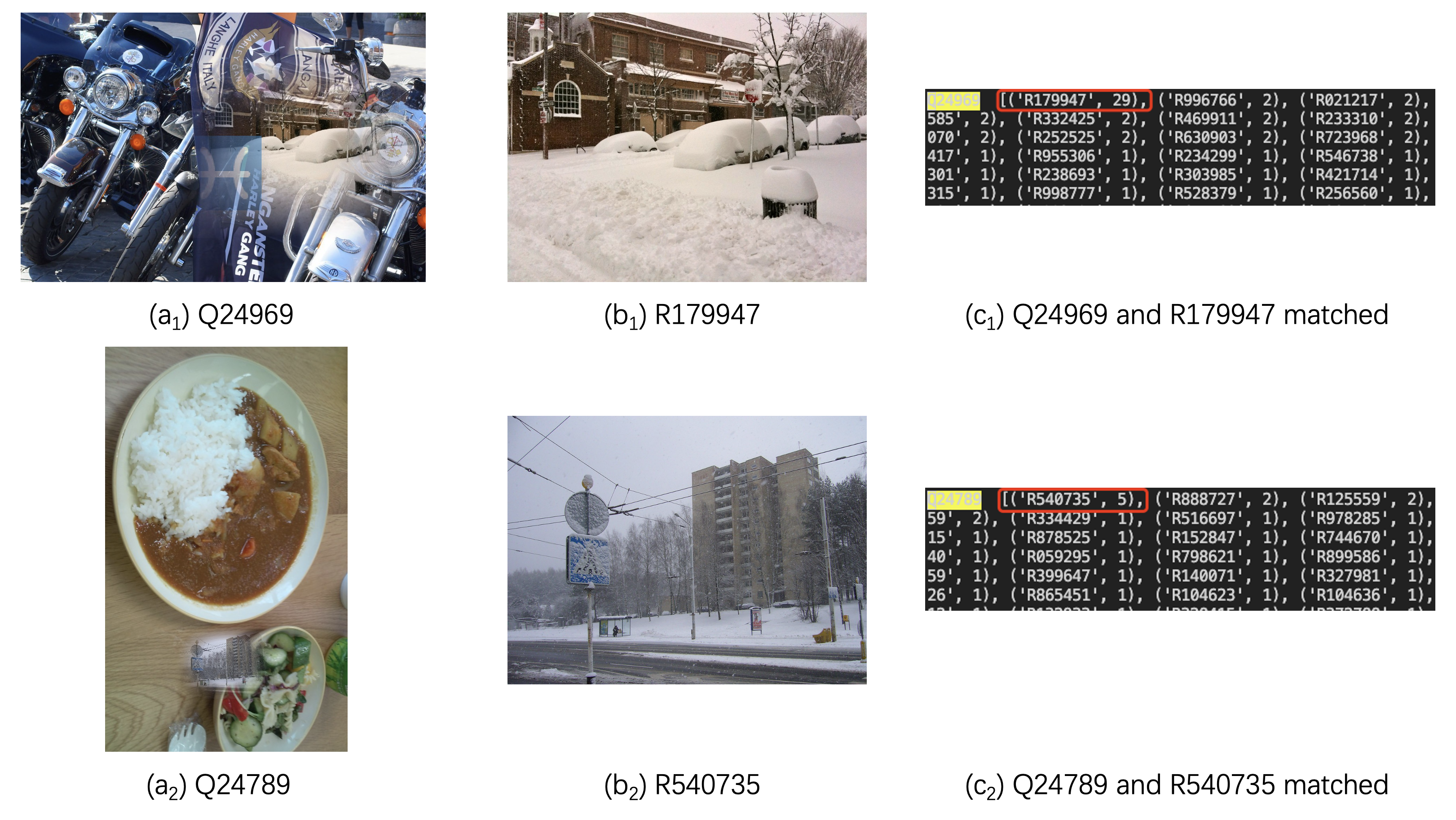}
\end{center}
  \caption{
  Hard case samples that local features can recall but global ones cannot. (a) represents query image, (b) represents reference image, (c) is the matching result, correspondingly.
  }
\label{fig:local_retrieval_sample}
\end{figure*}

 We find that global features still fail to recall some extreme cases, such as overlay on a background with a small part, as shown in the Fig. \ref{fig:AP_curve} and Fig. \ref{fig:local_retrieval_sample}. However, this kind of case has a strong degree of recognition in local features. Therefore, a very intuitive idea is to directly use local features for retrieval, so as to improve the recall rate.

~

\noindent\textbf{Local Feature Extract.}
  SIFT (Scale Invariant Feature Transform) is a classical and robust local feature extraction algorithm \cite{lowe1999object, lowe2004distinctive}.  It was proposed by David Lowe in 1999 and summarized in 2004. The SIFT features are invariant to image scaling, translation, rotation and  illumination changes, so we choose it as a local feature for recall.
  In practice, we firstly resize an image to a minimum edge of 300 pixels. And then generate some 128-dimensional vectors in uint8 type by SIFT. 

~

\noindent\textbf{GPU Faiss Retrieval.}
 Because there are about 600 million SIFT point features from 1 million reference images, we use GPU Faiss \cite{JDH17} to speed up the retrieval of local features. All SIFT features need about 165G GPU Memory for Float16 computing, so we run this process on a 8GPUs (Tesla V100) machine. After acceleration, it will take about 1-2 seconds for one query image during local feature recall.
 
  During retrieval, a query image will have multiple local features, each local feature will find the Top1 result using L2 distance. The results of all points in a query image are finally counted at the reference image level. Then L2 distance threshold and the number of matching points are used to judge the recall.
  
  We also try to use SuperPoint \cite{detone2018superpoint} instead of SIFT as a local feature for retrieval, but the test results are not ideal. We think it may be because SuperPoint is more suitable for matching tasks in combination with SuperGlue \cite{sarlin2020superglue}.

\subsection{Matching Score}
We use the matching point numbers of two images as matching score. As the same method (KNN-matching) proposed by Lowe with SIFT, we compare the distance between the nearest neighbor and the second-nearest neighbor to match the key points. When the ratio is less than the threshold (1/1.8), it can be considered as a correct matching pair and increasing the matching score by one. For a pair of matching images, we will flip the query image once to calculate the second matching score, and take the largest as matching result.

As described above, there are three recall branches: global recall, original local recall and  cropped local recall. We will add up the matching scores in the three branches as the final matching score. In addition, we think we can try to use SuperPoint+SuperGlue instead of SIFT+KNN to calculate the matching score in the future.

\section{Experiments}

\subsection{Training Settings}
Our model implementation is based on the Pytorch framework and Faiss library \cite{JDH17}.  And using 8 NVIDIA Tesla V100 for training the Transformer model. For training settings. We use Adamw \cite{loshchilov2018fixing} as the optimizer, the initial learning rate of model  finetune is set to 0.0001, and cosine scheduler \cite{loshchilov2016sgdr} is used to adjust the learning rate. For most models, we train 50 epochs on the training dataset, and train 200 epochs on the small number of labeled data in Phase 1. See our code for some details and differences. Besides, our local retrieval is accelerated by GPU and need a certain amount of memory to build the index database.

\subsection{Ablation Study}

The measure of effectiveness is micro-average precision ($\mu AP$) across all submitted image pairs, ranked by confidence score, also known as area under the precision-recall curve \cite{douze20212021}. See the official website for additional details. 

As shown in Table 1, we demonstrate the effectiveness of global model training pipline,  the Preprocessing Module, SIFT local retrieval. Note that the first finetune is self-supervised learning on the unlabeled training datasets, and the second finetune is on the half of the labels in Phase 1 (see details in Section 2.4), which will slightly overfit. Happily, our SIFT local model makes up for this part of the generalization ability.

In fact, there are still many competition skills that can be used, such as network model ensemble, network input/output size, layers hyper-parameter settings. Due to time constraints, we only tried some general methods. 

\begin{table}[t]
\centering
\resizebox{1.0\linewidth}{!}{
\begin{tabular}{lcc}  
\toprule
Model &Phase1 AP & Phase2 AP \\
\midrule
CNN G-emb (first F) &  0.335 & - \\
DINO G-emb (first F) & 0.568 & -\\
EsViT G-emb (first F) & 0.592 & - \\
EsViT G-emb (first F) + P & 0.612 & - \\
EsViT G-emb (second F) + P & 0.852 & 0.525\\
\midrule
SIFT &0.751  & - \\
EsViT G-emb (second F) +  P + SIFT & 0.898 & 0.768\\

\bottomrule\\
\end{tabular}
}
\label{table:ablation_study}
\caption{Results of ablation study. G-emb means global embedding, F means finetune, and P means using Preprocessing Module.}
\end{table}

\section{Conclusion}
In this paper, we proposed a global and local dual recall architecture for the Image Similarity Challenge (ISC) 2021 organized by Facebook AI, and won the 3rd place in matching track. In summary, we use data augmentations, memory bank,self-supervised learning and transformer backbone for training global features. And robust SIFT feature and GPU Faiss are used in local retrieval. Finally, the matching score is computed by KNN-matching. Our method reveals the complementary advantages of global and local features.

{\small
\bibliographystyle{ieee_fullname}
\bibliography{mybib}
}

\end{document}